\documentclass{bmvc2k}

\usepackage{times}
\usepackage{epsfig}
\usepackage{graphicx,subcaption}
\usepackage{amsmath}
\usepackage{amssymb}
\usepackage{gensymb}
\usepackage{xcolor}
\usepackage[normalem]{ulem}
\usepackage{verbatim}
\usepackage{algorithm,algorithmic}

\newcommand{\Skip}[1]{}

\newcommand{\YOON}[1]{
	\textcolor{blue}{
		\bfseries{YOON: {#1}}
	}
}

\newcommand{\CH}[1]{
	\textcolor{orange}{
		\bfseries{CH: {#1}}
	}
}

\newcommand*{\crossupsidedown}{%
    \text{%
      \raise .5ex\hbox{%
        \rlap{\vrule height.2pt depth.2pt width .75ex}%
        \hbox to .75ex{\hss\vrule height 1ex depth .5ex\hss}%
      }%
    }%
}

\title{In-N-Out: Towards Good Initialization for Inpainting and Outpainting}

\addauthor{Changho Jo}{timegate@kaist.ac.kr}{1}
\addauthor{Woobin Im}{iwbn@kaist.ac.kr}{1}
\addauthor{Sung-Eui Yoon}{sungeui@kaist.edu}{1}

\addinstitution{
	School of Computing,\\
	Korea Advanced Institute of Science and Technology (KAIST),\\ Daejeon, Korea\\
}

\runninghead{Jo et al.}{In-N-Out - Good Initialization for Inpainting and Outpainting}

\begin{document}

\maketitle

\begin{abstract}
In computer vision, recovering spatial information by filling in masked regions,
e.g., inpainting, has been widely investigated for its usability and wide applicability to 
other various applications: image inpainting, image extrapolation, and environment map estimation.
Most of them are studied separately depending on the applications. 
Our focus, however, is on accommodating the opposite task, e.g., image outpainting,
which would benefit the target applications, e.g., image inpainting.
Our self-supervision method, In-N-Out, is summarized as a training approach that leverages the knowledge of the opposite task into the target model.
We empirically show that In-N-Out 
-- which explores the complementary 
information --
effectively takes advantage over the 
traditional pipelines where
only task-specific learning takes place in training.
In experiments, 
we compare our method 
to the traditional procedure 
and analyze the effectiveness of our method on different applications: 
image inpainting, image extrapolation, and environment map estimation.
For these tasks, we demonstrate that In-N-Out consistently improves the performance of the recent works with In-N-Out self-supervision to their training procedure. Also, we show that our approach achieves better results than an existing training approach for outpainting. 
\end{abstract}

\section{Introduction}
\label{sec:intro}
\begin{figure}[t]
\centering
\begin{subfigure}[b]{0.49\textwidth}
\includegraphics[width=\linewidth]{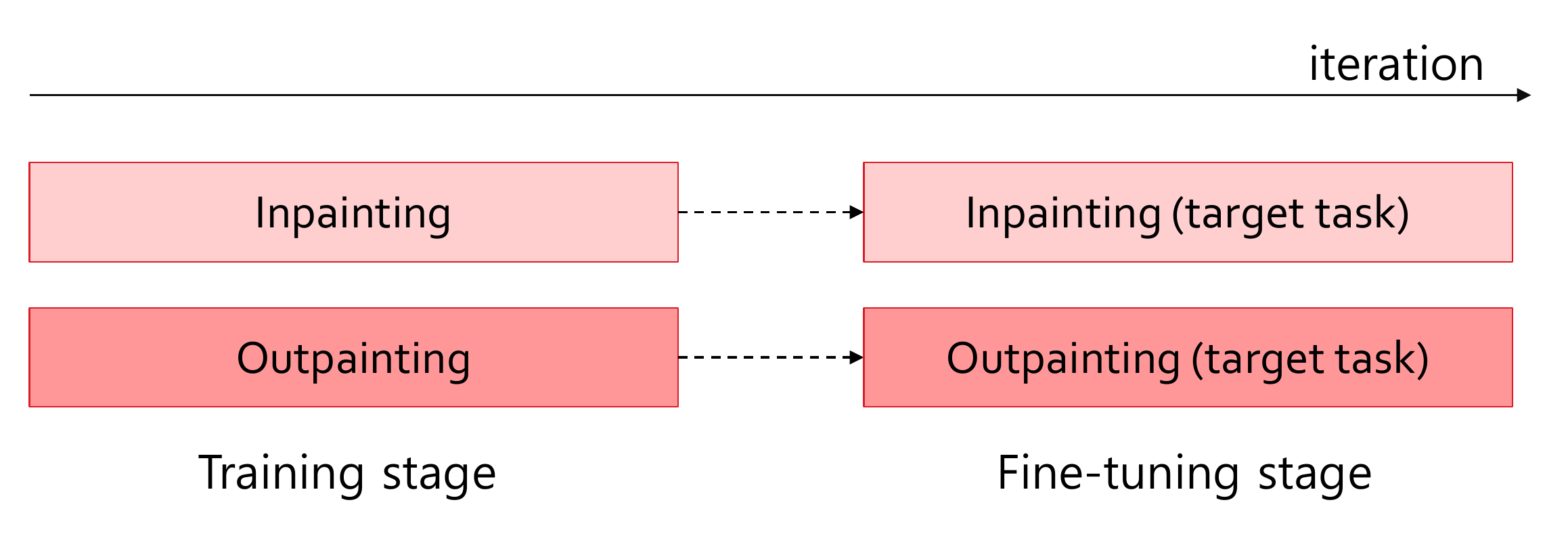}
\caption{Baseline}
\end{subfigure}
\hfill
\begin{subfigure}[b]{0.49\textwidth}
\includegraphics[width=\linewidth]{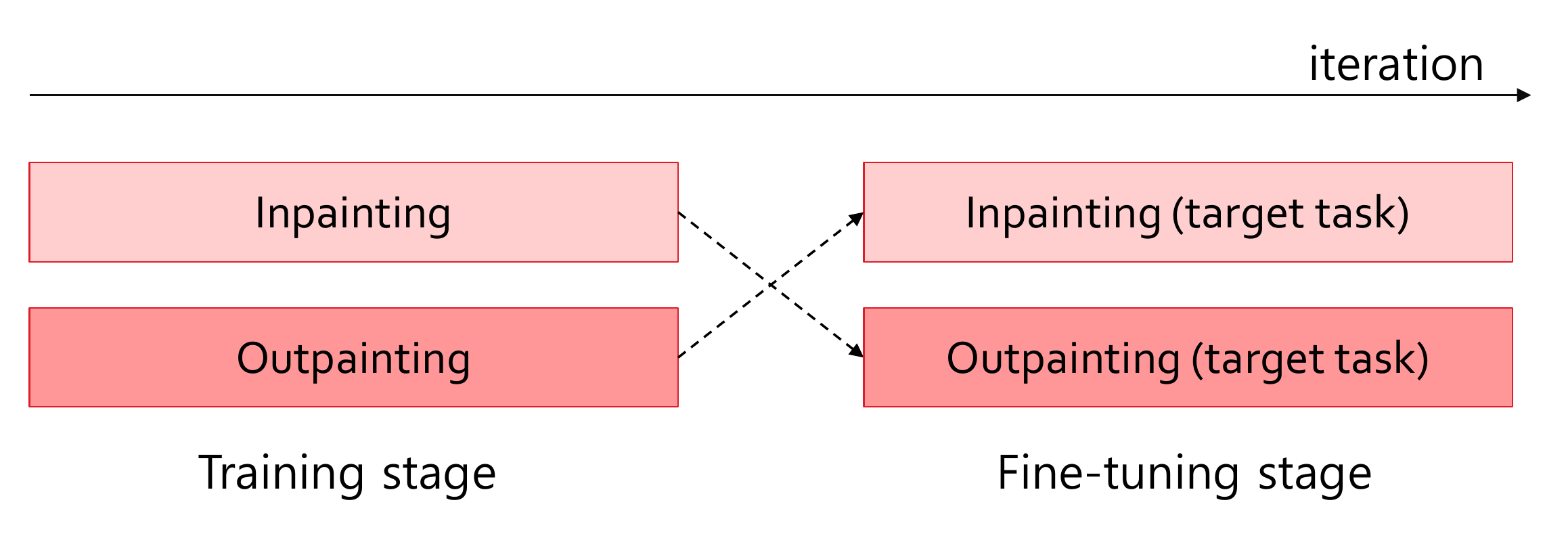}
\caption{In-N-Out}
\end{subfigure}
\caption{\textbf{(a)}: General procedure (baselines) for inpainting and outpainting, \textbf{(b)}: In-N-Out approach for inpainting and outpainting.}
\label{fig:inpainting_and_outpainting}
\end{figure}

\begin{figure}[t]
\centering
\begin{subfigure}[b]{0.23\linewidth}
\includegraphics[width=\linewidth]{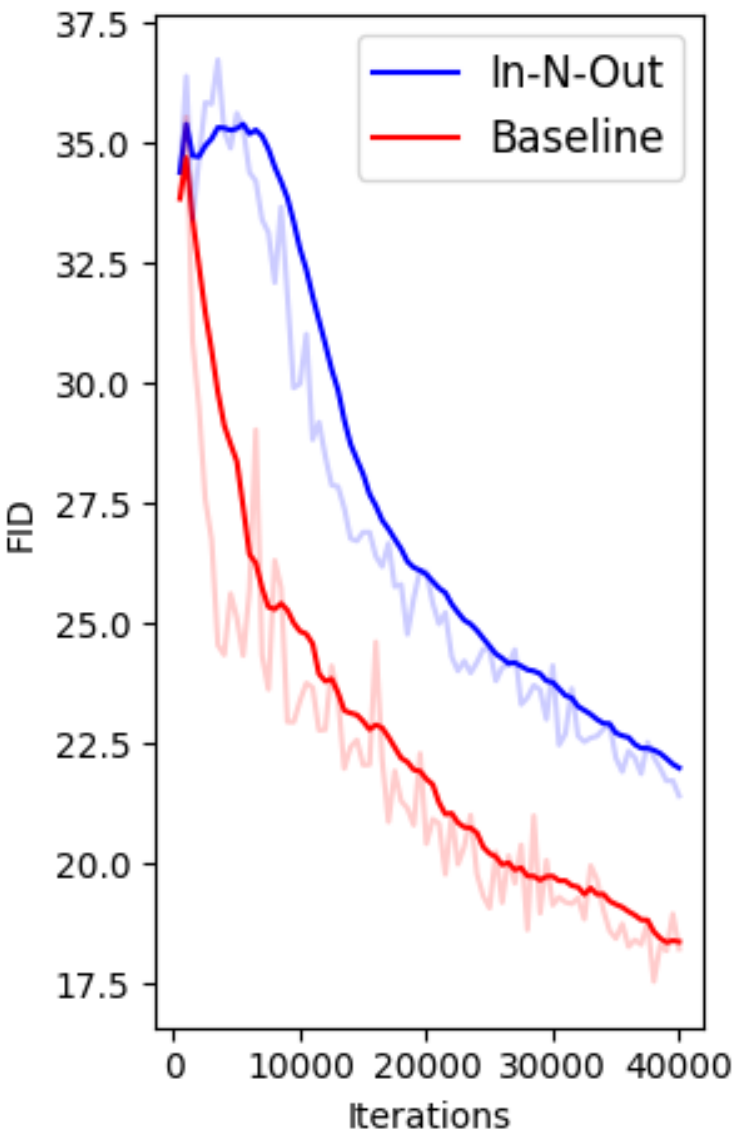}
\caption{Training}
\label{in_training_fid}
\end{subfigure}
\begin{subfigure}[b]{0.23\linewidth}
\includegraphics[width=\linewidth]{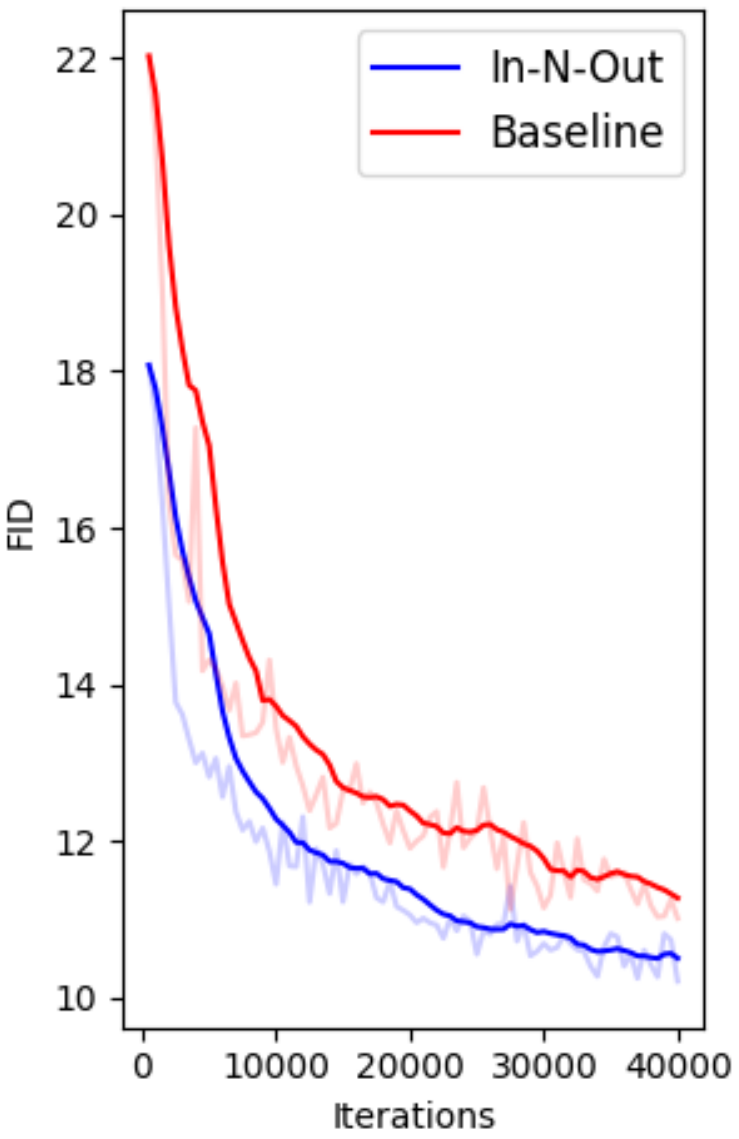}
\caption{Fine-tune from (a)}
\label{in_finetuning_fid}
\end{subfigure}
\begin{subfigure}[b]{0.23\linewidth}
\includegraphics[width=\linewidth]{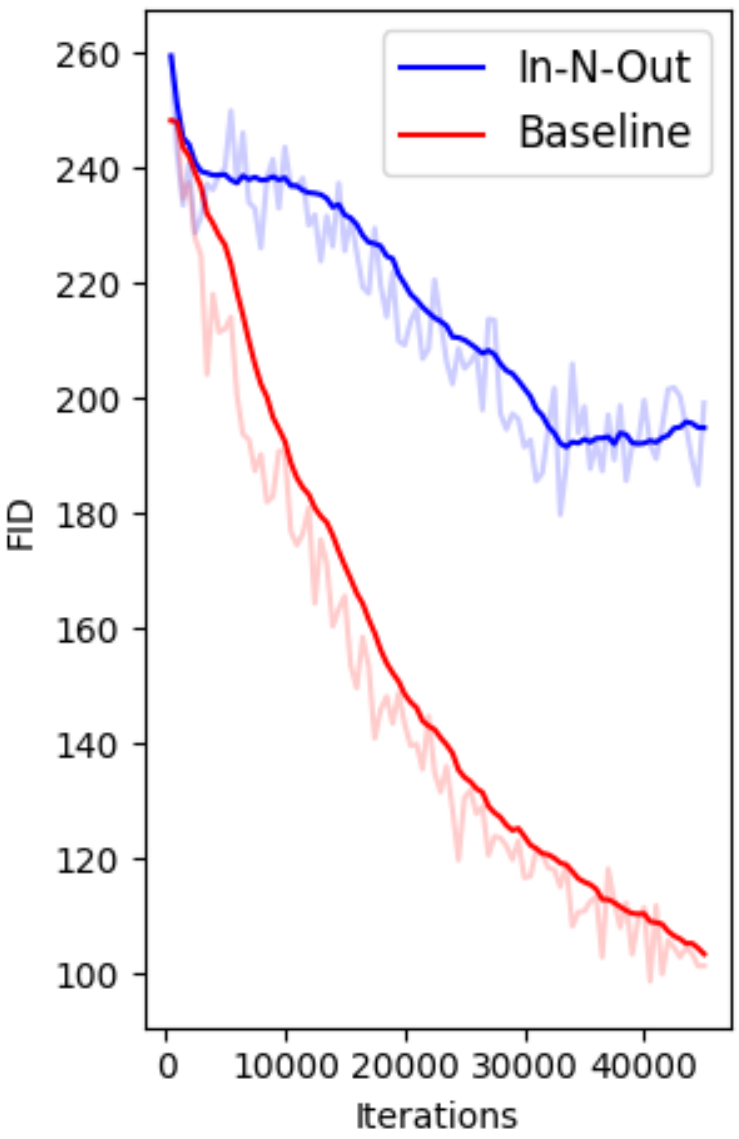}
\caption{Training}
\label{out_training_fid}
\end{subfigure}
\begin{subfigure}[b]{0.23\linewidth}
\includegraphics[width=\linewidth]{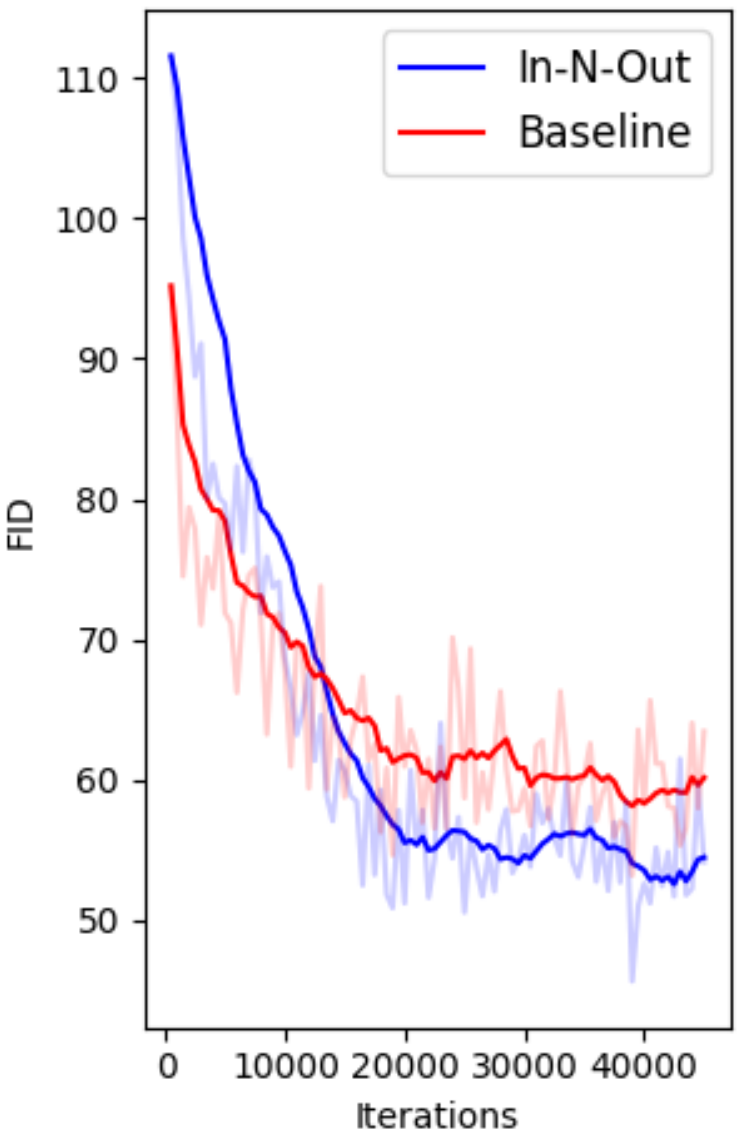}
\caption{Fine-tune from (c)}
\label{out_finetuning_fid}
\end{subfigure}

\caption{\textbf{(a)}-\textbf{(b)}: training procedure for image inpainting task ($\mathcal{T}_{\text{in}}$). 
Each graph represents \textbf{(a)} training stage from scratch, and  \textbf{(b)} fine-tuning stage.
\textbf{(c)}-\textbf{(d)}: training procedure for image outpainting task ($\mathcal{T}_{\text{out}}$). 
Each graph represents \textbf{(c)} training stage from scratch, and  \textbf{(d)} fine-tuning stage.
Each graph shows FID (Fréchet inception distance)~\cite{10.5555/3295222.3295408} during each stage.
Though our method, In-N-Out, 
shows worse performance in the training stage, 
In-N-Out converges more fastly and shows better performance in the end, where each method learns each target task by fine-tuning. 
Interestingly, good initialization does not require having a 
good metric score (FID) before fine-tuning. 
Also, it can be confirmed that In-N-Out (blue line) 
is an approach that can be well applied to various filling mask tasks such as inpainting and outpainting.}
\label{fig:full_fid}
\vspace{-0.2in}
\end{figure}

Learning to write a paper helps the learner
better fill in blanks in a sentence.
Learning to draw 
from scratch 
gives the ability to fill in the masks, i.e., inpainting.
Often, an educational curriculum for humans
gives learners 
such tasks, in the order we mentioned, or vice versa.

Thus, some tasks are complementary;
each task is mutually cooperative with each other
so that learning one task is eventually good for the other tasks.
So are tasks of computer vision,
which have been studied in the form of
transfer learning (pretraining)~\cite{wei2019semi, Guo_2019_CVPR}, multi-task learning~\cite{kendall2018multi, misra2016cross}, and continual learning~\cite{Zhou_2021_CVPR, zhai2019lifelong}.

In this work, we introduce a transfer learning
strategy, In-N-Out, for both image inpainting and image outpainting tasks.
In-N-Out is motivated by the complementary relationship 
between image inpainting and image outpainting tasks. 
\Skip{
Image inpainting task fills holes inside given image, and image outpainting extrapolate images
from borders of given image.
Although they are mainly developed separately, they commonly focused on context to carry deeper semantic understanding of the visible region to masked region
For example, to fill the deeper inside of masked region,~\cite{pathak2016context} leveraged {\it contextual} generative adversarial network. 
Also, outpainting networks adopt dilated convolution ~\cite{yang2019very, teterwak2019boundless,guo2019structure, wu2019deep, kasaraneni2020image, sabini2018painting}, and skip connection ~\cite{yang2019very,teterwak2019boundless,guo2019structure,zhang2020sienet} to fill the farther outer region. 
These similar developments imply that they can be complementary. 
}
For instance, changing an outpainting task to an inpainting task~\cite{kim2021painting} and 
cycle consistency between both tasks~\cite{kasaraneni2020image} are introduced. Inspired by this, 
we propose our method, In-N-Out, 
which is a simple and general training practice that can be applied to a network quickly.
In short, our method In-N-Out learns an outpainting task for the initialization of a model for an inpainting task, 
and learns an inpainting task 
for the initialization of a model for an outpainting task
as in Figure~\ref{fig:inpainting_and_outpainting}.

Figure~\ref{fig:full_fid} shows the test results of each training iteration. 
In Figure~\ref{in_training_fid} and Figure~\ref{out_training_fid}, 
although In-N-Out shows worse performance
as it does not learn the target task in the training stage, 
it shows the faster convergence and 
better performance than 
the baseline in the fine-tuning stage as shown in Figure~\ref{in_finetuning_fid} and Figure~\ref{out_finetuning_fid}.
Note that the baseline learns the target task 
in both the training and the fine-tuning stage, i.e., inpainting $\rightarrow$ inpainting for inpainting task.

\Skip{
When applying our approach to image inpainting, one possible explanation it works is to stimulate the network by giving them more difficult tasks (image outpainting). As Avramova ~\cite{avramova2015curriculum} showed that neural networks extract most learning values from the most difficult examples, our approach also can be seen as a way to grow given network strong, yet progressively. From this point of view, in that it gives stimulation with more difficult tasks, our approach have the spirit of hard example mining (HEM)~\cite{shrivastava2016training}.

On the other hand, when applying our approach to image outpainting, our approach can be interpreted as an extended work from cutout ~\cite{devries2017improved} or random erasing ~\cite{zhong2020random}, different in that our approach erases more region in fine-tuning stage and it is applied to image outpainting. As Chang et al. ~\cite{chang2017active} pointed out that emphasizing easier samples can lead to a better performance when the dataset is challenging or noisy, our approach can reinforce given network by sequentially training them from image inpainting task to image outpainting task. From this point of view, our approach has the same spirit of curriculum learning (CL) ~\cite{bengio2009curriculum}.
}

To summarize, our contributions are as follows:
\begin{itemize}
    \item This work provides a simple and general self-supervised pretraining method for both inpainting and outpainting tasks.
    \item We analyze and provide extensive experiments to validate
    In-N-Out procedure in both image inpainting and image outpainting tasks.
    \item We show the effectiveness of our method on various applications: inpainting task, image extrapolation, and environment map estimation.
\end{itemize}

\Skip{
For both image inpainting and image outpainting tasks, the most important part for performance is to synthesize the higher level features for masked region, from the deeper semantic understanding of the visible region. For example, to fill the more deeper inside of masked region, ~\cite{pathak2016context} leveraged {\it contextual} generative adversarial network. Also unlike inpainting networks, outpainting networks adopts dilated convolution ~\cite{yang2019very, teterwak2019boundless,guo2019structure, wu2019deep,kasaraneni2020image,sabini2018painting}, and skip connection ~\cite{yang2019very, teterwak2019boundless,guo2019structure, zhang2020sienet} to fill the more outer region.
}

\section{Related Work}
\label{sec:relwork}
\begin{figure}[t]
  \begin{subfigure}[b]{0.4\textwidth}
    \includegraphics[width=\linewidth]{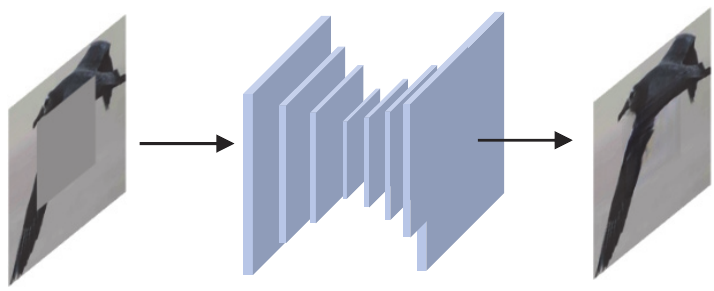}
    \caption{Inpainting pipeline}
    \label{fig:inpainting_pipeline}
  \end{subfigure}\hfill
  \begin{subfigure}[b]{0.4\textwidth}
    \includegraphics[width=\linewidth]{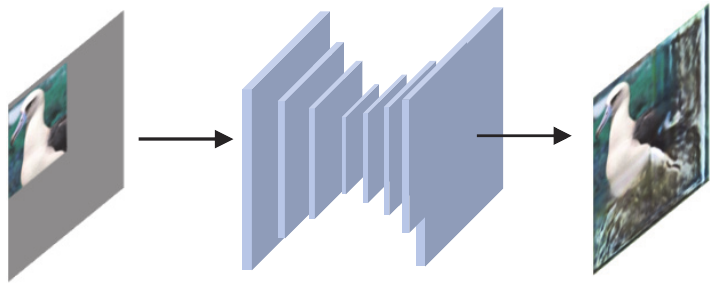}
    \caption{Outpainting pipeline}
    \label{fig:outpainting_pipeline}
  \end{subfigure}
  
  \caption{Image outpainting (or image extrapolation) shares the same pipeline as image inpainting, where the masked input image becomes the completed image. 
  }
  \label{fig:general_pipeline}
  \vspace{-0.2in}
\end{figure}

\paragraph{Deep Learning-based Inpainting and Outpainting.}
Image inpainting is aimed at filling holes in images. In a general image inpainting pipeline where the masked input image is filled in, generative adversarial networks (GANs)~\cite{goodfellow2014generative} are mainly employed, and various techniques have been developed in terms of network architectures~\cite{pathak2016context,yan2018shift,yu2018generative} and loss functions~\cite{pathak2016context,yang2017high,iizuka2017globally,mechrez2018contextual,wang2018image} to infer
the invisible masked region better from visible known regions.
New layers have been proposed to transfer textures from input images, including shift-connection layer~\cite{yan2018shift} and contextural attention layer~\cite{yu2018generative}.
In addition, loss functions have been proposed in terms of global and local consistency~\cite{yang2017high,iizuka2017globally}, and
patch distribution~\cite{mechrez2018contextual,wang2018image}.

On the other hand, image outpainting 
fills outer masked regions from visible inner regions. 
Similar to the image inpainting task, image outpainting has been developed mainly in terms of network architectures~\cite{yang2019very,wang2019wide} and loss functions~\cite{yang2019very,wang2019wide,teterwak2019boundless}.
Outer propagation of inner visible features has been proposed with a recurrent module~\cite{yang2019very} and a normalization module~\cite{wang2019wide}.
WGAN~\cite{gulrajani2017improved} losses are mainly used for outpainting methods~\cite{yang2019very, wang2019wide, teterwak2019boundless} and spatially variant losses are applied~\cite{wang2019wide}.

Apart from the architecture and losses, we focus on the
complementary relationship between both tasks:
can knowledge from one side give an advantage to the other side?
Regarding that both tasks do not require a specific label, i.e., self-supervised, exploiting the opposite task is feasible and desirable.
It has been shown that exploring across the tasks is beneficial by 
solving outpainting with inpainting~\cite{kim2021painting}, or
learning cycle-consistency between both tasks~\cite{kasaraneni2020image}.
We step further by simplifying the cross-task learning 
and generalize it to both tasks
with transfer learning, which is a well-known general strategy in deep learning~\cite{zamir2018taskonomy}.

Specifically, our method (In-N-Out) lets a model first experience 
the opposite task (e.g., inpainting), 
which is transferred to the target task (e.g., outpainting).
We show by the experiment (Table~\ref{table:beach_finetuned_results}) that our simple transfer strategy
gives an equivalent or a larger amount of performance gain 
to the task-specific (i.e., horizontal extrapolation) method~\cite{kim2021painting}, which also exploits the cross-task relation.
Compared to common practice for the inpainting or outpainting, 
where pretraining on the target task is performed,
our method gives no overhead, since it substitutes the pretraining task with 
the opposite task without additional overhead.



\paragraph{Self-supervised Learning.}
Self-supervised learning has been proposed to learn feature representations from annotation-free images.
For instance, image transformation~\cite{dosovitskiy2015discriminative}, rotation \cite{komodakis2018unsupervised}, position of paches~\cite{doersch2015unsupervised}, and color~\cite{zhang2016colorful} can be used as surrogate labels obtained from the image itself for feature learning.
While a majority of self-supervision methods is validated on classification tasks, 
different kinds of features fit better for different tasks~\cite{zamir2018taskonomy}.
Therefore, specific self-supervision methods have been proposed for better performance in different tasks:
optical flow~\cite{liu2019selflow,im2020unsupervised}, 
reflection removal~\cite{Yang_2018_ECCV,kim2020single},
image inpainting~\cite{pathak2016context, wang2018image}, image outpainting~\cite{sabini2018painting, teterwak2019boundless},
and light estimation~\cite{10.1145/3130800.3130891, song2019neural}.

Our method is regarded as a self-supervised method,
which is defined by performing the opposite task (e.g., inpainting) for the target task (e.g., outpainting).
Our self-supervision is devised to help the `filling masked region' task,
including inpainting, extrapolation, and environment map estimation.

\section{In-N-Out: Inpainting and Outpainting}\label{sec:method}
\begin{figure}[t]
\centering
  \includegraphics[width=\linewidth]{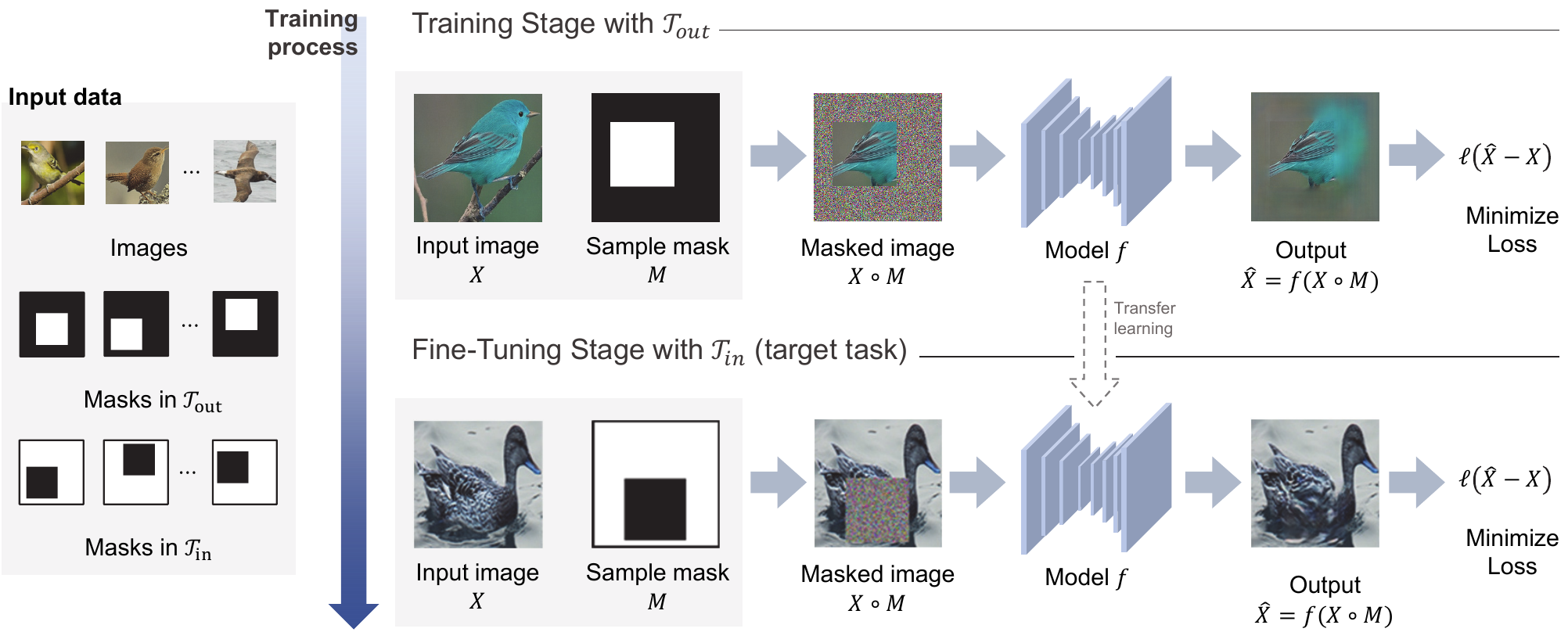}
  \caption{In-N-Out training process for inpainting task.}
  \label{fig:training_process}
  \vspace{-0.2in}
\end{figure}

In this section, we propose a training scheme, In-N-Out, for the inpainting and outpainting tasks.
The key idea of our method is to enable the knowledge from a counterpart 
task; the counterpart task is defined as inpainting 
if the main task is outpainting, and outpainting if the target task is inpainting (Fig.~\ref{fig:inpainting_and_outpainting}).
We found that transfer learning, which has been widely applied and investigated by computer vision researchers, 
shows a fair amount of effectiveness in our purpose.
Thus, we divide the whole training process into two parts: \textbf{training} with the counterpart and \textbf{fine-tuning} with the target task (Fig.~\ref{fig:inpainting_and_outpainting}).

Specifically, let $\mathcal{T}$ denote a task, which is a data distribution for the general `fill masked region' task.
That is, $\mathcal{T}$ is a joint distribution of images and masks,
where we sample image $X$ and mask $M$ from distribution $\mathcal{T}$ to perform a self-supervision task;
the task is to learn to restore $X$ from the input that masked by mask $M$.
We denote
inpainting task $\mathcal{T}_{\text{in}}$ 
and 
outpainting task $\mathcal{T}_{\text{out}}$.
Inpainting task $(X,M)\sim\mathcal{T}_{in}$ generates sample mask $M$ whose masked region is local, which is often called as a hole (or holes) in an image (Fig.~\ref{fig:inpainting_pipeline}). 
On the other hand,
outpainting mask $M$ from distribution $\mathcal{T}_{out}$ has the visible region locally, 
and the outer regions are masked (Fig.~\ref{fig:outpainting_pipeline}).
For each task, a prediction model $f_\theta$ can be trained by minimizing loss:
\begin{equation}
    \mathcal{L}(\theta) =
    \mathbb{E}_{(X, M)\sim\mathcal{T}}\left[ \ell(f_\theta(X \circ M), X) \right],
\label{eq:total_loss}
\end{equation}
where $\circ$ is the Hadamard product operator with filling operation for the masked region; the region is filled with normally uniform value~\cite{yeh2017semantic, pathak2016context} or random noise~\cite{li2017generative, deng2018uv}.
Note that $\ell(A,B)$ is a loss function to measure a distance between $A$ and $B$, e.g., L1 loss.
Our loss includes the adversarial loss, ID-MRF loss~\cite{wang2018image}, etc., depending on the target tasks; the details are mentioned as implementation details in Section~\ref{sec:expr}.

We transfer knowledge from opposite tasks ($\mathcal{T}_{\text{out}}$ and $\mathcal{T}_{\text{in}}$), and  define our process as follows.
If the target task is $\mathcal{T}_{\text{in}}$, 
we run the first $N$ steps with $\mathcal{T}_{\text{out}}$ with loss function 
$\mathcal{L}(\theta)$ in Eq.~\ref{eq:total_loss}.
Then, we continue to fine-tune on the target task $\mathcal{T}_\text{in}$
for $K$ steps additional to the training step.
This simple rule applies vice versa when $\mathcal{T}_{\text{out}}$ is the target.
In short,
$\mathcal{T}_{\text{out}} \rightarrow \mathcal{T}_{\text{in}}$ for inpainting task, and
$\mathcal{T}_{\text{in}} \rightarrow \mathcal{T}_{\text{out}}$ for outpainting task.

Intuition for transferability
is based on the complementary property 
between inpainting and outpainting.
Inpainting and outpainting have been taken to account together~\cite{kim2021painting, kasaraneni2020image}.
We simplify and show the relationship in the form of transfer learning.
We empirically show that this simple approach
can have an equivalent or better performance than the task-specific
method~\cite{kim2021painting}.

\paragraph{Transfer learning.} 
It has been shown that visual tasks help each other~\cite{zamir2018taskonomy}, e.g., learning surface normal $\rightarrow$ learning depth, 
which is effectively shown with transfer learning.
Our work aims to discover the relationship between inpainting and outpainting with transfer learning, and it provides an effective yet simple methodology for an inpainting or an outpainting task.

\paragraph{Complementarity.} 
Inpainting fills a hole region by seamlessly expanding a visible outer area into the hole 
and outpainting transfers a context from 
a small visible region to an invisible outer region.
The difference between the hidden regions -- a hole or an outer area --
makes an inpainting model focus on low-level restoration of the hole and
an outpainting network focus on deeper understanding of context, 
rather than low-level prediction.
Thus, learning from both tasks lets the model have
a sort of synergy by benefiting from both high and low level knowledge. 
Similarly, in literature, an outpainting task is often regarded as 
a more difficult one than inpainting from the perspective of a model.
Regarding this, In-N-Out can be viewed as 
curriculum learning~\cite{bengio2009curriculum} or reverse curriculum learning~\cite{wang-etal-2019-dynamically}.

\section{Experiments \& Results}
\label{sec:expr}
We evaluate our method and show its generality on various applications in both inpainting and outpainting tasks.
The evaluation tasks include image inpainting, image outpainting, and environment map estimation.
Throughout all tasks, we follow the scheme described in Sec.~\ref{sec:method} and Fig.~\ref{fig:inpainting_and_outpainting}.
That is, for an outpainting task (e.g., environment map estimation), 
we use the inpainting task before fine-tuning. 
We start a model from scratch with the self-supervised training,
followed by the fine-tuning step for each desired task.
For all experiments, we use a variant of Semantic Regeneration Network~\cite{wang2019wide},
unless otherwise specified. We provide the architecture details in the supplementary material.
Also, for inpainting and outpainting, 
many types of masks have been used. 
For instance, uniform values (to black or white), or noises are used. 
We use uniform noise that randomly chooses all possible color values
unless otherwise stated.
In the subsequent subsections, we describe the evaluation results of each task.

\subsection{Image Inpainting}
\label{section:ablation_study_inpainting}
\begin{table}[t]
\centering
\begin{subtable}[b]{0.49\textwidth}
\centering
\begin{tabular}{lccc}
\hline
Method                   & PSNR           & SSIM           & FID            \\ \hline
Baseline                 & 21.03          & 0.830          & 11.02          \\
\textbf{In-N-Out} & \textbf{21.14} & \textbf{0.833} & \textbf{10.22} \\ \hline
\end{tabular}
\caption{Image inpainting task.}
\label{table:inpainting_finetuned_results}
\end{subtable}
\hfill
\begin{subtable}[b]{0.49\textwidth}
\centering
\begin{tabular}{lccc}
\hline
Method                   & PSNR           & SSIM           & FID            \\ \hline
Baseline                 & \textbf{15.11}          & 0.552          & 63.45         \\ 
\textbf{In-N-Out} & 15.10 & \textbf{0.554} & \textbf{54.13} \\ \hline
\end{tabular}
\caption{Image outpainting task. 
}
\label{table:outpainting_finetuned_results}
\end{subtable}

\begin{subtable}[b]{0.49\textwidth}
\centering
\begin{tabular}{lccc}
\hline
Method & PSNR & SSIM & FID \\
\hline
SRN~\cite{wang2019wide} & 18.22 & 0.513 & - \\
PSL~\cite{kim2021painting} & 18.78 & {\textbf{0.716}} & 35.86 \\
\textbf{In-N-Out} & {\textbf{19.52}} & 0.711 & {\textbf{30.17}} \\
\hline
\end{tabular}
\caption{Outpainting compared to PSL~\cite{kim2021painting}.}
\label{table:beach_finetuned_results}
\end{subtable}
\begin{subtable}[b]{0.49\linewidth}
\centering
\begin{tabular}{lccc}
\noalign{\smallskip}\noalign{\smallskip}\hline
Method & PSNR & SSIM & FID \\
\hline
Baseline & 24.83 & 0.813 & 63.63 \\
\textbf{In-N-Out} & {\textbf{24.86}} & {\textbf{0.817}} & {\textbf{59.91}} \\
\hline
\end{tabular}
\caption{Inpainting using irregular masks~\cite{yu2019free}, and MEDFE~\cite{Liu2019MEDFE} network.}
\label{table:MEDFE_finetuned_results}
\end{subtable}
\begin{subtable}[b]{0.49\linewidth}
\centering
\begin{tabular}{lccc}
\noalign{\smallskip}\noalign{\smallskip}\hline
Method & PSNR & SSIM & FID \\
\hline
Baseline & 31.64 & 0.950 & 29.63 \\
\textbf{In-N-Out} & {\textbf{32.26}} & {\textbf{0.955}} & {\textbf{22.46}} \\
\hline
\end{tabular}
\caption{Inpainting using irregular masks~\cite{liu2018image}, and Shift-Net~\cite{yan2018shift}.}
\label{table:ShiftNet_finetuned_results}
\end{subtable}

\caption{
\textbf{(a)} Results of In-N-Out (outpainting $\rightarrow$ inpainting) and baseline (inpainting $\rightarrow$ inpainting) for the inpainting task. 
\textbf{(b)} Results of In-N-Out (inpainting $\rightarrow$ outpainting) and baseline (outpainting $\rightarrow$ outpainting) for the outpainting task.
\textbf{(c)} Results of In-N-Out compared to PSL~\cite{kim2021painting}, for the outpainting task.
\textbf{(d)} Results of In-N-Out and baseline for the inpainting task, using irregular masks~\cite{yu2019free} on MEDFE network.
\textbf{(e)} Results of In-N-Out and baseline for the inpainting task, using irregular masks~\cite{liu2018image} on Shift-Net.
In-N-Out can be a good practice by leveraging the knowledge from the opposite task.
}

\label{table:inpainting_and_outpainting_table}
\end{table}

\begin{figure}[t]
\centering
\begin{subfigure}[b]{0.49\textwidth}
\centering
\includegraphics[width=\linewidth]{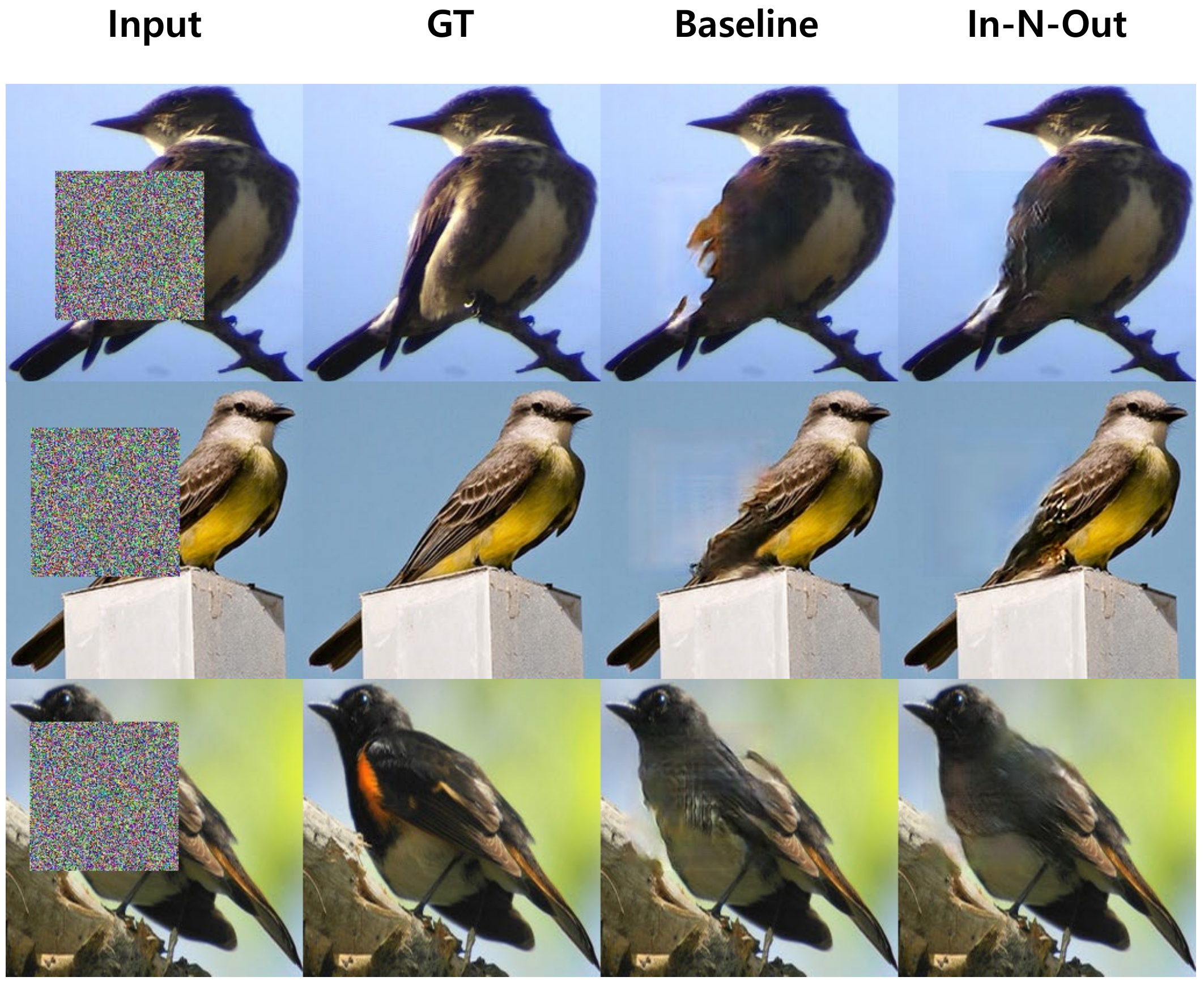}
  \caption{Inpainting task.}
  \label{fig:inpainting_imgs}
\end{subfigure}
\begin{subfigure}[b]{0.49\textwidth}
\centering
\includegraphics[width=\linewidth]{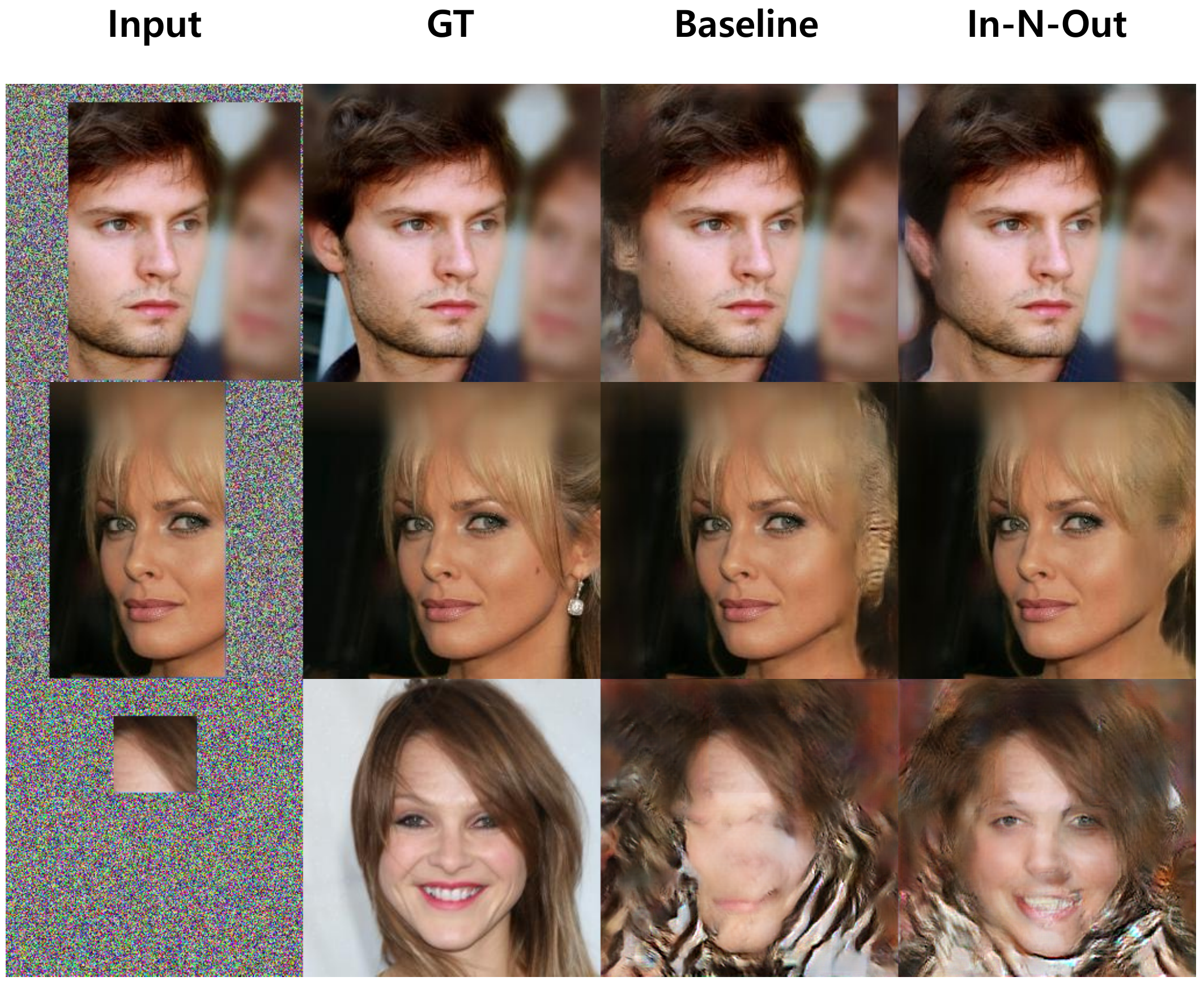}
  \caption{Outpainting task.}
  \label{fig:outpainting_imgs}
\end{subfigure}
\caption{Qualitative results on inpainting task (CUB200 dataset~\cite{welinder2010caltech}) and outpainting task (CelebA-HQ dataset~\cite{karras2017progressive}).}
\vspace{-0.2in}
\end{figure}

\paragraph{Experimental Setting.}
For the inpainting task, we tested our models on the CUB200 dataset~\cite{welinder2010caltech}, consisting of 11,788 bird images. 
We randomly crop the images with given bird locations and resize the images 
into $256\times 256$. For the size of the random mask, i.e., blocked region, 
we use resolution $128 \times 128$. 
The training and test splits are the same as in ~\cite{wang2019wide}.
For In-N-Out, the model goes through the schedule of outpainting $\rightarrow$ inpainting (Fig.~\ref{fig:training_process}).
On the other hand, the baseline for comparison has the schedule that 
has the same training scheme to the fine-tuning step, 
i.e., inpainting $\rightarrow$ inpainting, 
which has been a customary process for many computer vision tasks. 
For each stage, we use the strategies in ~\cite{wang2019wide}, which is 
training with reconstruction losses, 
then adding the adversarial loss and ID-MRF loss~\cite{wang2018image} in the fine-tuning stage. We run 40,000 iterations of training steps and 40,000 iterations of fine-tuning steps with batch size 8.
In the training stage, baseline uses $128\times 128$ mask 
and learns to paints inside the region,
while In-N-Out uses inverse mask and learns to paint outside of the $128\times 128$ region. 
Both baseline and In-N-Out learn to inpaint in the fine-tuning stage.

\paragraph{Quantitative Results.}
We provide image inpainting results 
on CUB200 dataset 
in Table~\ref{table:inpainting_finetuned_results}. 
When compared to the baseline, improvements can be observed in all three metrics: PSNR, SSIM, and Frechet Inception Distance (FID)~\cite{10.5555/3295222.3295408}. 
While the amount of improvements are not significant in PSNR and SSIM,
FID is shown to be improved 
in a clear gap (11.02 $\rightarrow$ 10.22).
That is because FID measures the quality of 
generated images, which is considered more important in image inpainting;
PSNR or SSIM can possibly be worse even
with higher quality images as in also noted by \cite{wang2019wide}.
This performance improvement is supported by
the training graph (Fig.~\ref{in_training_fid} and \ref{in_finetuning_fid}).
Even though In-N-Out shows higher FID
in the initial training stage (Fig.~\ref{in_training_fid}), 
In-N-Out
converges faster in the fine-tuning stage (Fig.~\ref{in_finetuning_fid}),
since it gives better initialization than the baseline.

\paragraph{Qualitative Results.}
The visual comparisons are shown in Figure~\ref{fig:inpainting_imgs}. Compared to the baseline, In-N-Out shows more convincing results. In-N-Out better restores the overall shapes of the birds -- including their wings and bodies -- with reasonable color tones,
where the baseline often fails to create continuous shape and suffers from some artifacts. More results can be found in our supplementary material.

\subsection{Image Outpainting}
\label{section:ablation_study_outpainting}
\paragraph{Experimental Setting.}
For the outpainting task, we test the models on
CelebA-HQ dataset~\cite{karras2017progressive} which consists of 30,000 face images. 
To show resiliency of our approach to the mask size, we use rectangular masks with random sizes for our testing.
The details of used masks are described in the supplementary material. 
We use the official training and test split of the dataset, 
and we resize images to $256 \times 256$ for both training and testing.
We train with reconstruction losses, then adversarial loss and ID-MRF loss are added in the fine-tuning stage as in Section~\ref{section:ablation_study_inpainting}.
With batch size 8, we run 45,000 iterations of training steps and 45,000 iterations of fine-tuning.
In the training stage, the baseline learns to paint outside of a $128 \times 128$ inner region, while In-N-Out learns to inpaint. 
Both baseline and In-N-Out learn to outpaint in the fine-tuning stage.

\paragraph{Quantitative Results.}
Test FIDs are provided in Figure~\ref{out_training_fid} and \ref{out_finetuning_fid}, and the numerical results are reported in Table~\ref{table:outpainting_finetuned_results}. As shown in Table~\ref{table:outpainting_finetuned_results}, In-N-Out shows better results in terms of FID. 
Interestingly, In-N-Out shows to be more effective on the outpainting task than on the inpainting task (Table~\ref{table:inpainting_finetuned_results}). This is partly because, inpainting tends to be learned easier than outpainting, since pixels to be inpainted can get more directional cues~\cite{kim2021painting}, so that the GAN is able to be stabilized faster.
This is also observed in the graph (Fig.~\ref{out_finetuning_fid}), where 
the baseline FID measure doesn't improve much in the fine-tuning stage, even with its 
lower error in the training stage (Fig.~\ref{out_training_fid}).
This is also supported by the results reported by
the original SRN paper~\cite{wang2019wide},
where 
quantitative measures get worsen\footnote{Please refer to Table 5 in \cite{wang2019wide} for more details.} 
after finetuning on SRN, which is our baseline network. 
On the other hand, 
In-N-Out is shown to ameliorate the issue (Fig.~\ref{out_finetuning_fid}),
and shows better quantitative results after fine-tuning.


\paragraph{Qualitative Results.}
We also provide visual comparisons in Figure~\ref{fig:outpainting_imgs}. In-N-Out shows plausible restoration of hair, mouth, and ear. While baseline suffers from some artifacts to far invisible region from the mask, In-N-Out shows plausible images to the far invisible region. More results are available in our supplementary material.

\Skip{
By these ablation studies (Section~\ref{section:ablation_study_inpainting} and Section~\ref{section:ablation_study_outpainting}), we validate that In-N-Out is a practically better procedure for both the inpainting task and outpainting task. Truly, it tells us that the knowledge from the opposite task is essentially needed for faster convergence and better performance. 
}

\begin{figure}[t]
\centering
\begin{subfigure}[b]{0.49\textwidth}
\centering
\includegraphics[width=\linewidth]{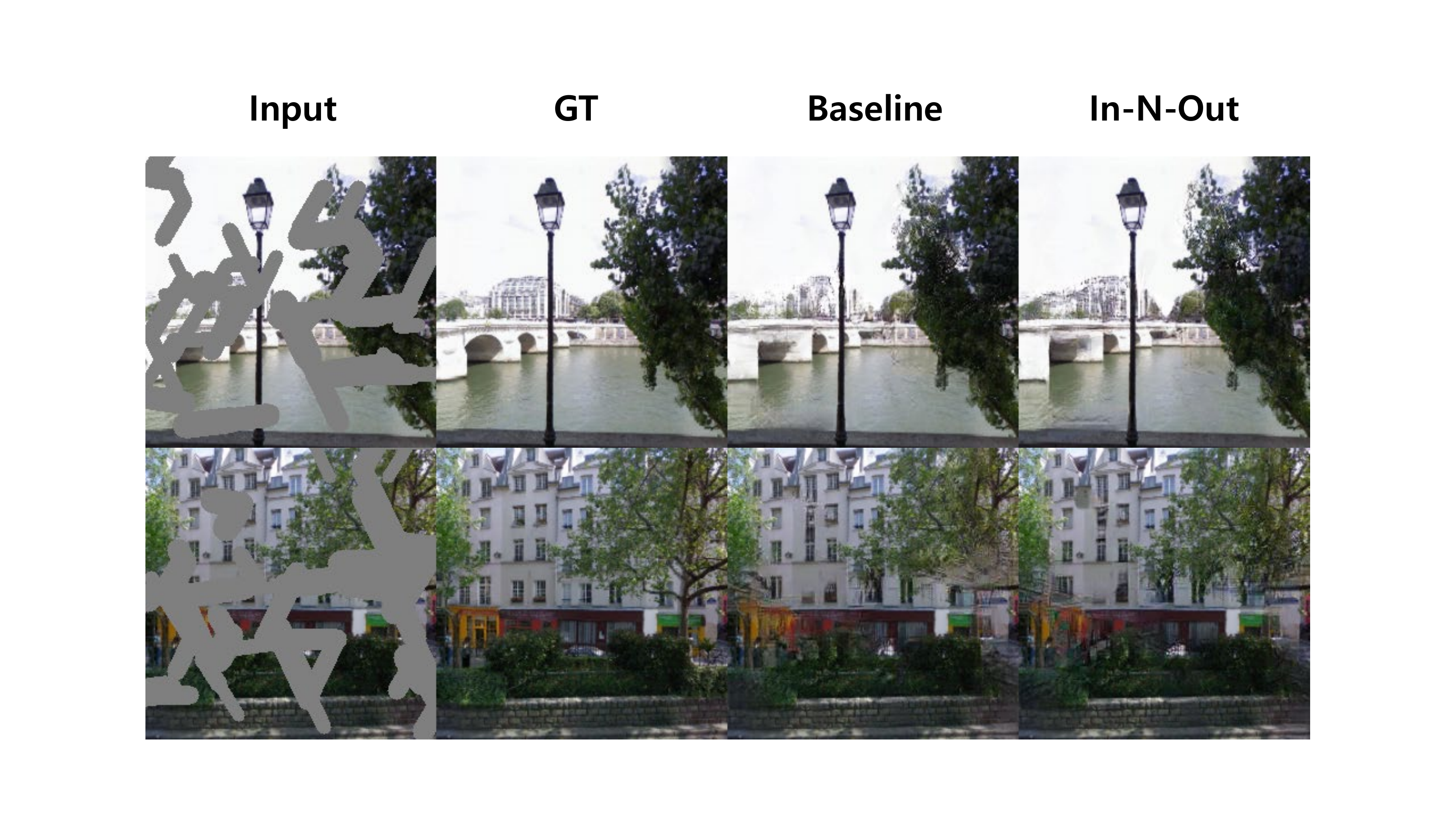}
  \caption{Inpainting task using irregular masks~\cite{yu2019free}.}
  \label{fig:MEDFE_imgs}
\end{subfigure}
\begin{subfigure}[b]{0.49\textwidth}
\centering
\includegraphics[width=\linewidth]{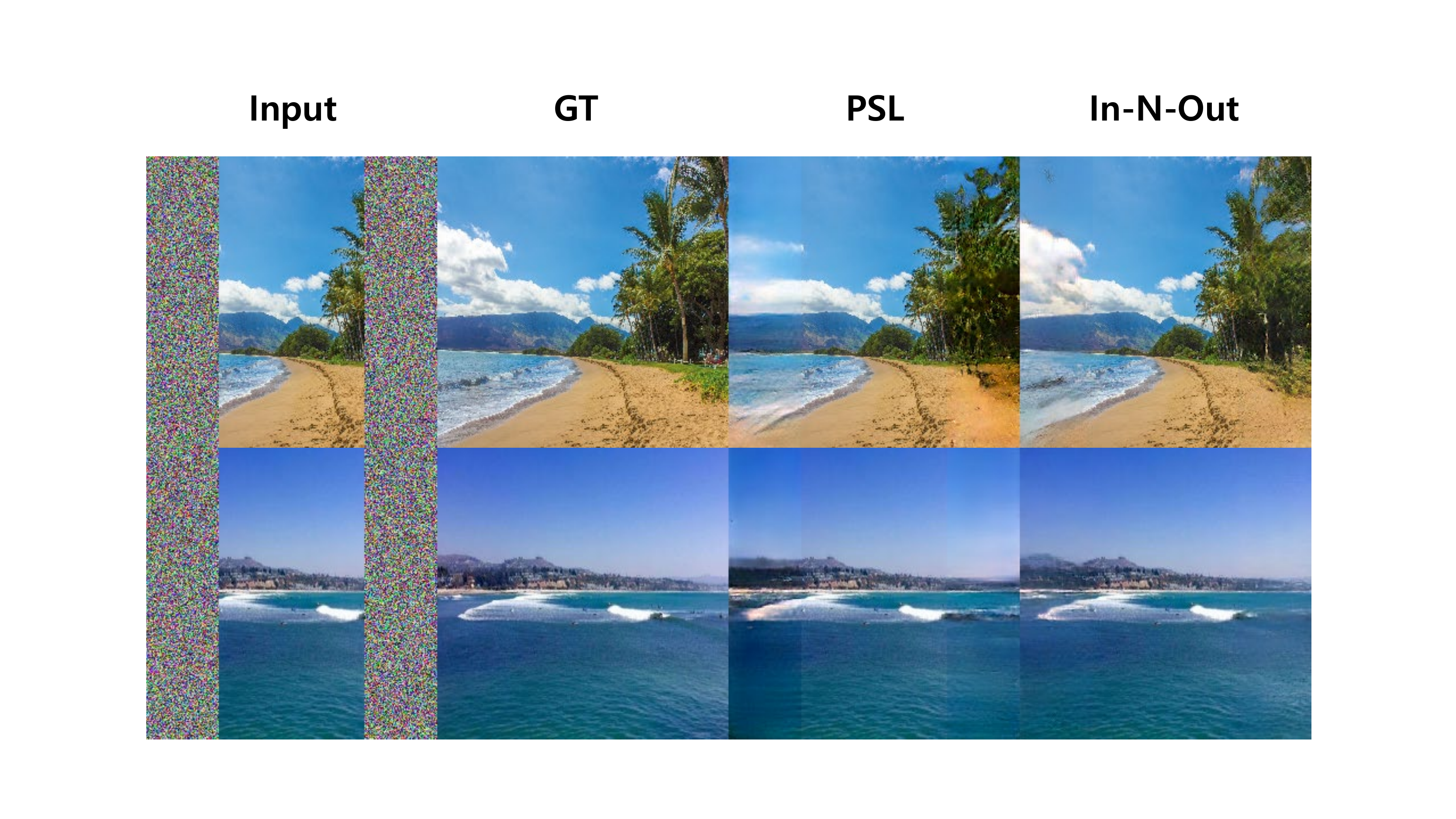}
  \caption{Outpainting task compared to PSL~\cite{kim2021painting}.}
  \label{fig:outpainting_beach_imgs}
\end{subfigure}
\begin{subfigure}[b]{0.49\textwidth}
\centering
\includegraphics[width=\linewidth]{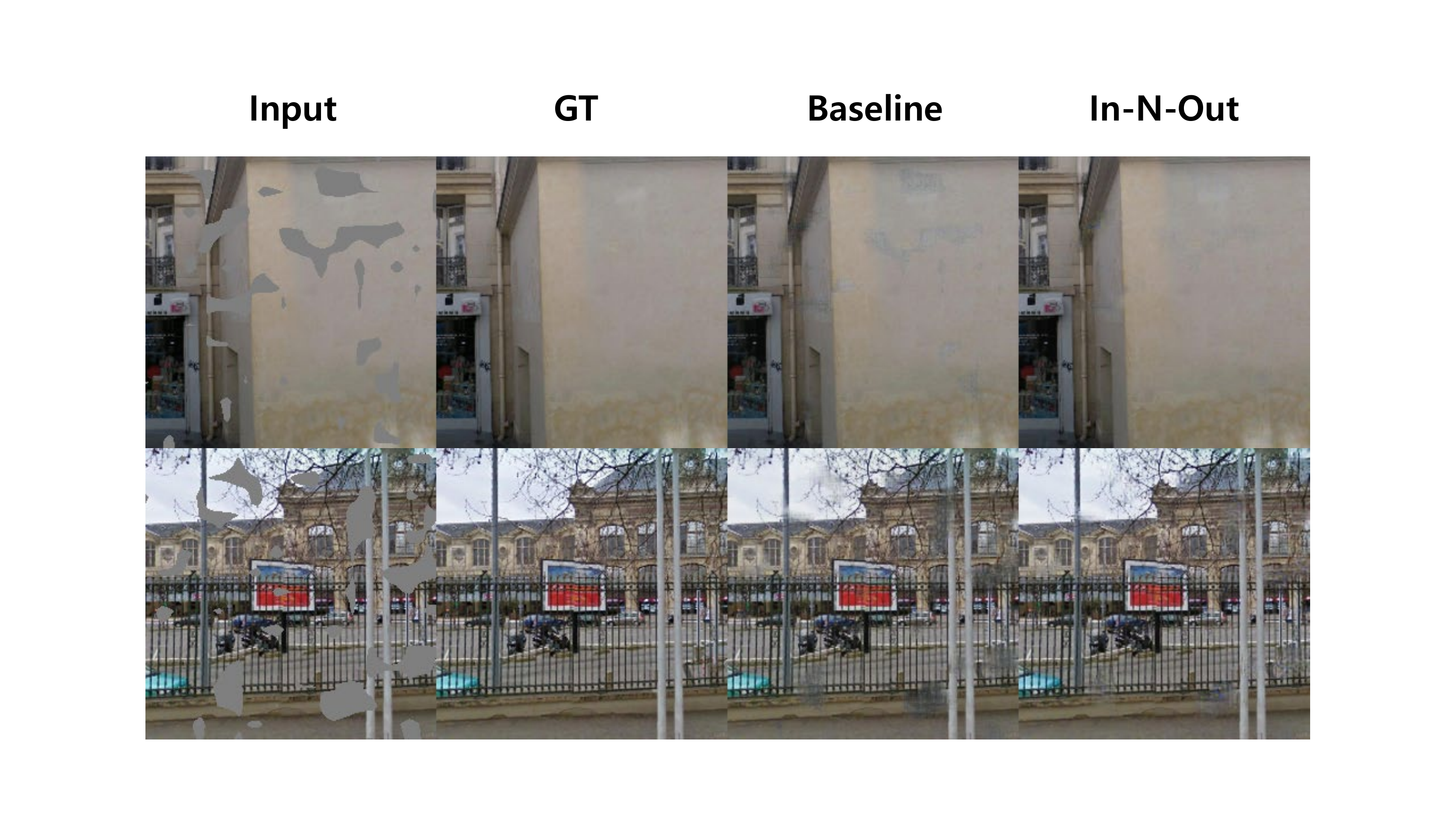}
  \caption{Inpainting task using irregular masks~\cite{liu2018image}.}
  \label{fig:ShiftNet_imgs}
\end{subfigure}
\caption{Qualitative results on inpainting task (Paris Street-View dataset~\cite{doersch2012what}) and outpainting task (beach dataset~\cite{sabini2018painting}). In-N-Out shows more seamless and semantically plausible results.
}
\end{figure}

\subsection{Comparison to PSL~\cite{kim2021painting}}
\label{subsection:beach}
\paragraph{Experimental Setting.}
We provide evaluation compared to progressive step learning (PSL)~\cite{kim2021painting}, which is a strategy where an outpainting task is 
intentionally substituted by an inpainting task.
We compare our method with PSL, 
since PSL gives knowledge from the opposite task (i.e., inpainting) for the target task (i.e., outpainting), similar to In-N-Out. In short, PSL changes the outpainting task to the inpainting task and performs progressive inpainting.
Please refer to \cite{kim2021painting} for details.
Compared to PSL, our method is simple and 
general, since we do not constrain our method to a specific task.
To compare with the approach, 
we train the model on the beach dataset~\cite{sabini2018painting},
which contains 10,515 images of nature landscapes.
With batch size 8, we run 40,000 iterations of training steps and 40,000 iterations of fine-tuning steps.

\paragraph{Results.}
In Table~\ref{table:beach_finetuned_results},
we report the performance of our baseline network (SRN), which is originally reported in \cite{kim2021painting}, progressive step learning (PSL)~\cite{kim2021painting},
and In-N-Out (ours).
First, we notice that In-N-Out clearly improves
over the baseline network (SRN) on PSNR and SSIM metrics.
When compared to a task-specific method (PSL),
In-N-Out shows clearly better results in PSNR and FID,
while SSIM is in a similar range.
As mentioned in the earlier sections, 
FID has been regarded as more important than PSNR or SSIM in generation tasks,
since it better measures the quality of generated images.
In Figure~\ref{fig:outpainting_beach_imgs}, In-N-Out predicts seamless (e.g., color of sky) and more naturally extrapolated scenery (e.g., clouds) than PSL.

\subsection{Image Inpainting with Irregular Masks}
\label{subsection:irregular_mask}
\paragraph{Experimental Setting.}
We also test our method on inpainting tasks with irregular masks, from  DeepFillv2~\cite{yu2019free} and PConv~\cite{liu2018image}. 
We use Paris Street-View dataset~\cite{doersch2012what}, which consists of 15,000 street-view images. We use masks with a uniform constant and networks designed for inpainting, i.e., MEDFE~\cite{Liu2019MEDFE} and Shift-Net~\cite{yan2018shift}. The details of experiment setting are available in our supplementary material.

\paragraph{Results.}
We provide quantitative results in Table~\ref{table:MEDFE_finetuned_results} and Table~\ref{table:ShiftNet_finetuned_results}.
In-N-Out shows improvements from the baseline in terms of FID. 
Also, in Figure~\ref{fig:MEDFE_imgs} and Figure~\ref{fig:ShiftNet_imgs}, 
it shows to predict background structures (first row of~\ref{fig:MEDFE_imgs}) and shape of trees (second row of~\ref{fig:MEDFE_imgs} and~\ref{fig:ShiftNet_imgs}) better. 
These experiments demonstrate that our In-N-Out is widely applicable to different shapes of masks including irregular masks,
and different network architectures.

\subsection{Environment Map Estimation}
\label{subsection:env_map}
\begin{figure}[t]
\centering
  \includegraphics[width=0.9\linewidth]{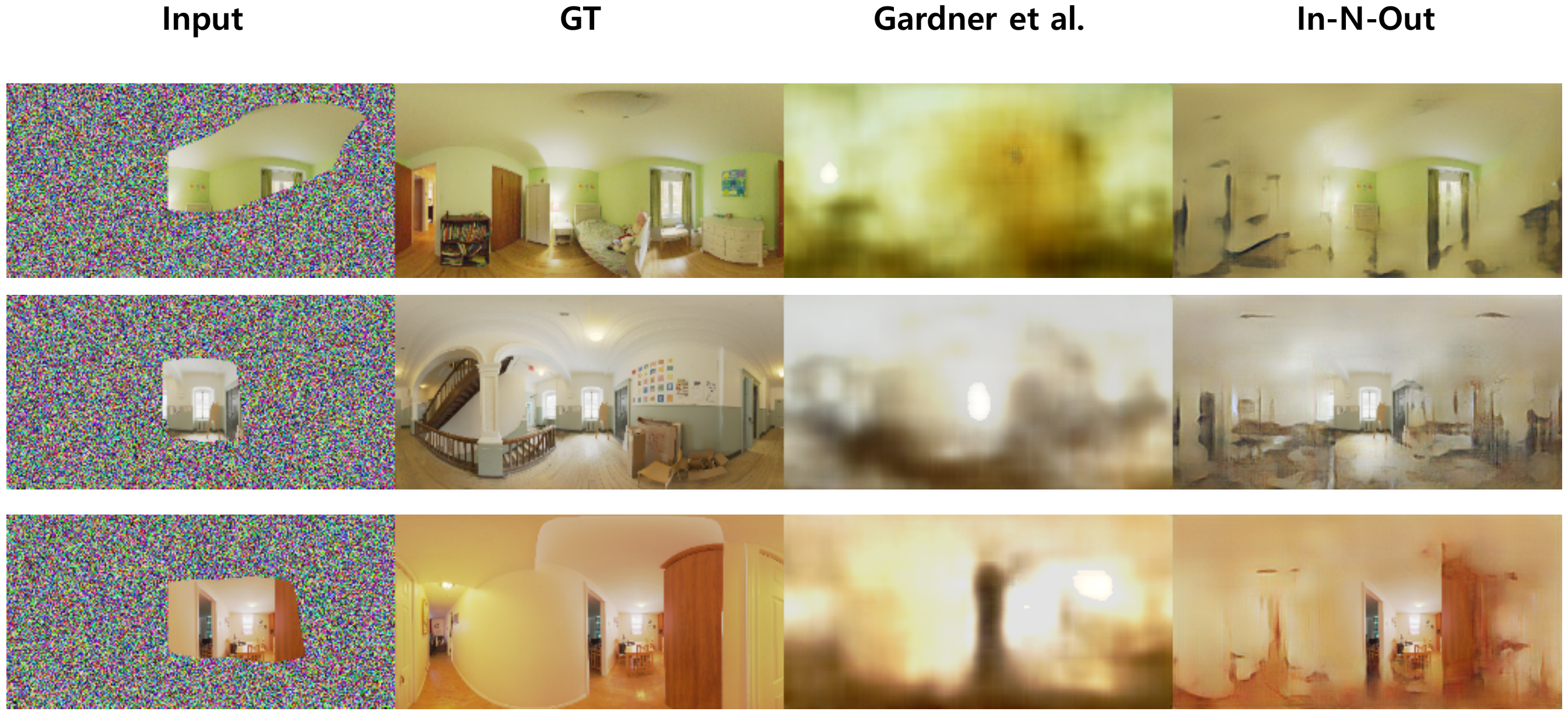}
  \caption{Visual comparison on environment map estimation task. The results demonstrate that our In-N-Out approach can be useful for light estimation or have more plausible tone-mapped results. These results show that our In-N-Out can help another application, environment map estimation.}
  \label{fig:envmap_imgs}
\end{figure}

\paragraph{Experimental Setting.}
Our approach also finds its application to environment map estimation. To get a supervision from HDR images, we added LDR to HDR conversion network from ~\cite{gkitsas2020deep} and train our model in an end-to-end manner. 
We train and test our model on the Laval Indoor HDR Dataset~\cite{10.1145/3130800.3130891} 
that consists of 2,233 indoor panoramas. 
The details of experiment setting and test masks are available in our supplementary material.

\paragraph{Qualitative Results.}
We provide results compared to Gardner {\it et al.},
even though it uses a different mask configuration to ours~\footnote{Masks correspond to the central 60-degree fov, and output panoramas are post-processed with the spherical warping.},
since the result gives an insight about the benefit of In-N-Out for
the environment map task. 
In detail, In-N-Out successfully recovers reflection of light in the ceiling (first row), overall color tones in the panorama (second row), and the shape of the floor (third row). While Gardner {\it et al.} is predicting well in terms of light, but it may fail in semantic detail (e.g., overall color-tone of the ceiling and floor).

\section{Conclusion}
By leveraging mutual complementarity, we have proposed 
In-N-Out, a training strategy for both inpainting and outpainting, which 
can work with various networks and loss functions.
In-N-Out learns an outpainting task for the initialization of a model for an inpainting task, and vice versa.
We showed they could benefit each other, and our approach shows promising results in inpainting, outpainting, and environment map estimation.

Although we showed the effectiveness of our approach, there is more interesting future work lies ahead.
We tackled the two tasks: inpainting and outpainting, while there are numerous `fill in masks' tasks between them. We expect we can devise a better self-supervision task from one side to others by a smoother transition with these tasks.
Moreover, 
we would like to point out that
our work can get waterfall effects from developments of network architectures
because our approach is parallel to the network architectures. For example, leveraging transferability of Learning without Forgetting (LwF)~\cite{li2017learning} 
may improve our approach further. 
Also, context encoders~\cite{pathak2016context} showed that pretrained features using image inpainting can help other computer vision tasks (e.g., classification, detection, and semantic segmentation); our work may be able to help other vision tasks by leveraging inpainting and outpainting, and showing this is left as future work.

\bibliography{egbib}
\end{document}